\documentclass[letterpaper, 10 pt, conference]{ieeeconf}
\usepackage{times}
\usepackage{graphicx}
\usepackage{amsmath,amssymb,amsopn,amstext,amsfonts}
\usepackage{cancel}
\usepackage[space]{cite}
\usepackage{pdfsync}
\usepackage{balance}
\usepackage{color}
\usepackage{mathtools}
\usepackage{algpseudocode}
\usepackage{algorithm} 
\usepackage{bm}

\usepackage{diagbox}
\usepackage{float}
\usepackage{epstopdf}
\usepackage{pifont}
\usepackage{fixltx2e}
\usepackage{amsmath}
\usepackage{multirow}
\usepackage{url}
\usepackage{verbatim}
\makeatletter
\let\NAT@parse\undefined
\makeatother
\usepackage[linkcolor=black,citecolor=black,urlcolor=black,colorlinks=TRUE]{hyperref}
\usepackage{booktabs}
\usepackage{graphicx}
\usepackage{subcaption}
\usepackage{gensymb}
\graphicspath{{}}
\DeclareGraphicsExtensions{.png,.jpg,.eps,.pdf}
\IEEEoverridecommandlockouts
\begin{document}
	\title{\LARGE \bf Fast-Tracker 2.0: Improving Autonomy of Aerial Tracking with Active Vision and Human Location Regression}
	\author{Neng Pan, Ruibin Zhang, Tiankai Yang, Chao Xu, and Fei Gao 	
	\thanks{All authors are with the State Key Laboratory of Industrial Control Technology, Zhejiang University, Hangzhou 310027, China, and also with the Huzhou Institute of Zhejiang University, Huzhou 313000, China.  \tt \small  \{panneng\_zju, ruibin\_zhang, tkyang, cxu, and fgaoaa\}@zju.edu.cn}}
	
	\maketitle
	\thispagestyle{empty}
	\pagestyle{empty}
	\begin{abstract}
	\label{sec:abstract}\textbf{}
	In recent years, several progressive works\cite{TN2017elli, Bonatti2019JFR, BJ2020ICRA, JC2016tracking, HZP2020tracker} promote the development of aerial tracking.
	One of the representative works is our previous work Fast-tracker\cite{HZP2020tracker} which is applicable to various challenging tracking scenarios. However, it suffers from two main drawbacks: 1) the over simplification in target detection by using artificial markers and 2) the contradiction between simultaneous target and environment perception with limited onboard vision.
	 In this paper, we upgrade the target detection in Fast-tracker\cite{HZP2020tracker} to detect and localize a human target based on deep learning and non-linear regression to solve the former problem. For the latter one, we equip the quadrotor system with 360\degree active vision on a customized gimbal camera. 
	 Furthermore, we improve the tracking trajectory planning in Fast-tracker\cite{HZP2020tracker} by incorporating an occlusion-aware mechanism that generates observable tracking trajectories. Comprehensive real-world tests confirm the proposed system's robustness and real-time capability. Benchmark comparisons with Fast-tracker\cite{HZP2020tracker} validate that the proposed system presents better tracking performance even when performing more difficult tracking tasks.
	 
	
	\end{abstract} 

	\IEEEpeerreviewmaketitle
	
	\section{Introduction}
	Tracking a moving target with flying robots blooms in both industry and academia in recent years. 
	Thanks to the advancement of vision and navigation technology, aerial tracking has shown significant robustness and impressive agility.
	In our previous work, we present a complete aerial tracking system Fast-Tracker~\cite{HZP2020tracker}, which enables a quadrotor to autonomously track an agile target with uncertain intention in unknown cluttered environments, and has been successfully applied in versatile scenarios.
	
\begin{figure}[t]
		\centering
		\begin{subfigure}{1\linewidth}
			\centering
			\includegraphics[width=1\linewidth]{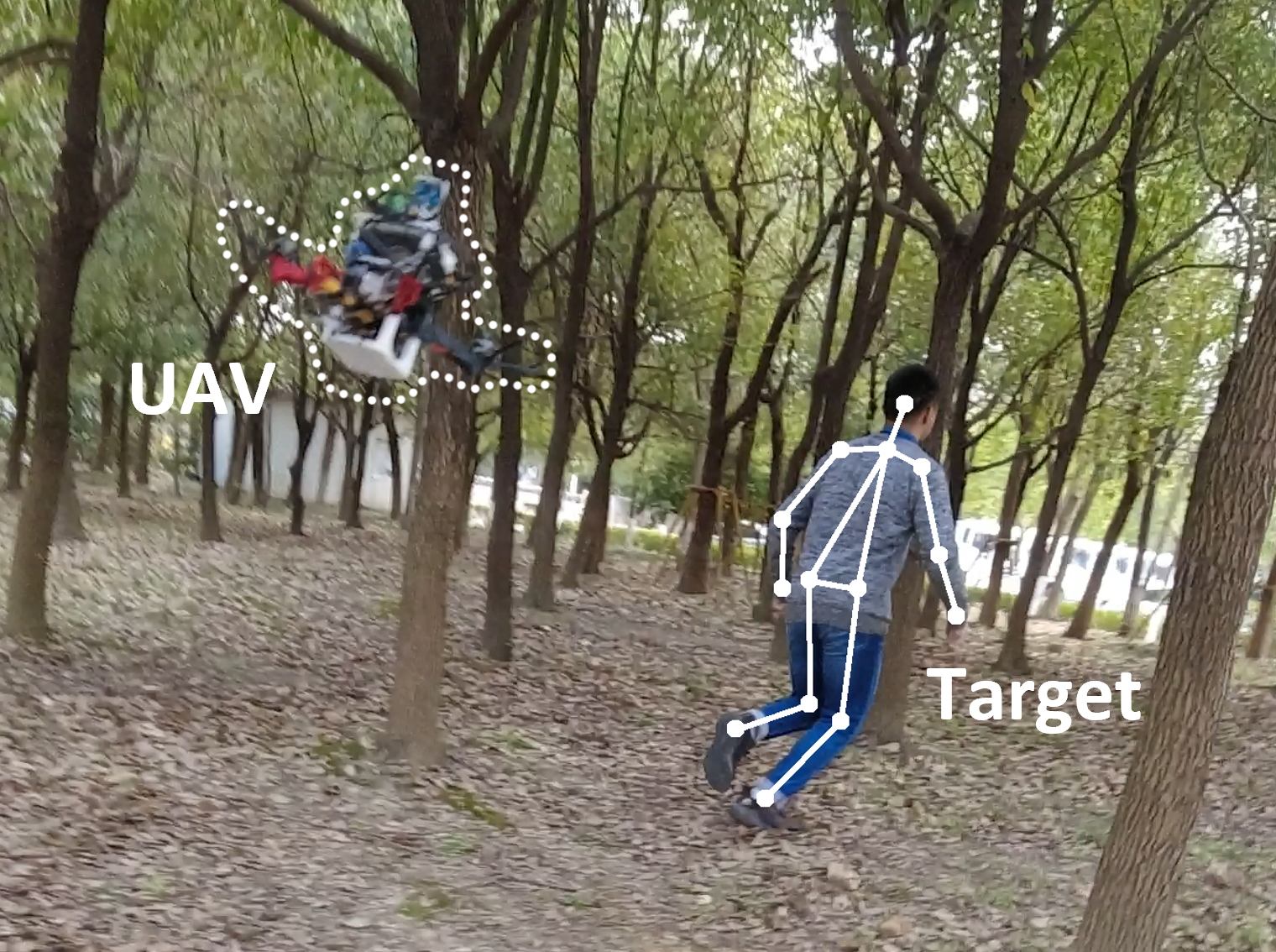}
			\captionsetup{font={small}}
			\caption{The outdoor experiment in a dense forest.}
			\label{pic:outdoor_experiment}
		\end{subfigure}
		\begin{subfigure}{1\linewidth}
			\centering
			\includegraphics[width=1\linewidth]{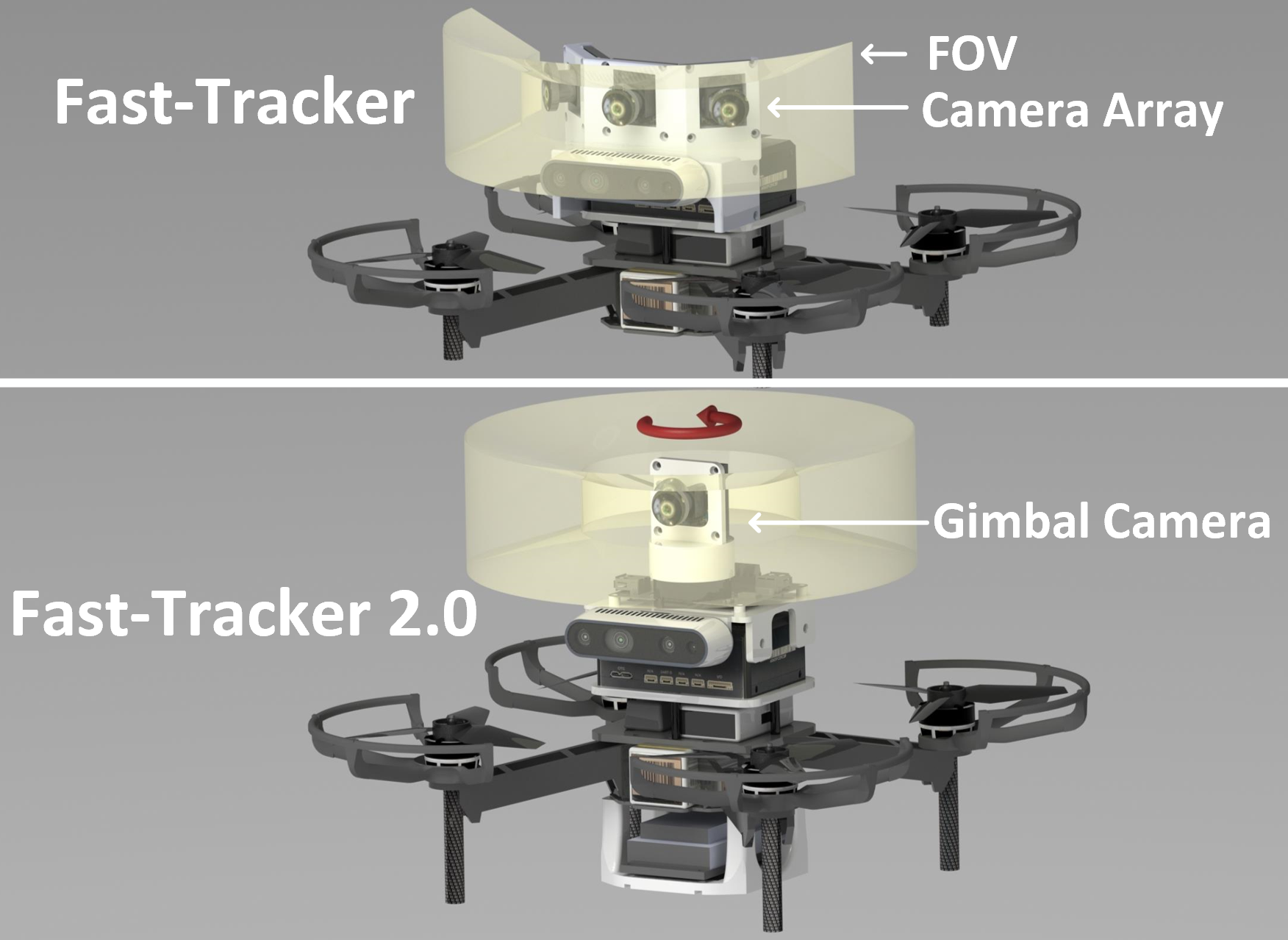}
			\captionsetup{font={small}}
			\caption{Comparison of the hardware design between the proposed and the previous~\cite{HZP2020tracker} quadrotor systems. Compared with the fixed FOV in Fast-Tracker~\cite{HZP2020tracker}, the proposed system enables free rotation of the camera. }			
			\label{pic:comparison}
		\end{subfigure}
		\captionsetup{font={small}}
		\caption{
			Demonstration of the outdoor experimental scenario and the proposed hardware improvement.
		}
		\vspace{-1.1cm}
		\label{pic:top_graph}
	\end{figure}

	However, Fast-Tracker~\cite{HZP2020tracker} still has severe drawbacks. 
	Chiefly, the artificial marker-based target detection module hinders the wider application of our system.
	In this paper, we upgrade the target perception as a marker-less one, which detects a human target based on deep learning and estimates its location with pre-trained non-linear regression. 
	Besides, as pointed out in Fast-Tracker~\cite{HZP2020tracker}, it is contradictory to simultaneously detect the target and surrounding obstacles with the limited field of view (FoV). 	
	To resolve this, we replace the camera array mechanism in Fast-Tracker~\cite{HZP2020tracker} with a lightweight gimbal onboard the quadrotor (as shown in Fig.\ref{pic:comparison}) to separate the vision for tracking and for environment sensing. 
	The conflict between the target and environment has been resolved with the actively controllable gimbal camera, and the quadrotor itself focuses more on safe navigation.
	In addition, we incorporate an occlusion penalty in our path searching method that encourages the quadrotor to find observable paths while tracking.
	
	\begin{figure*}[t]
		\centering
		\includegraphics[width=1\linewidth]{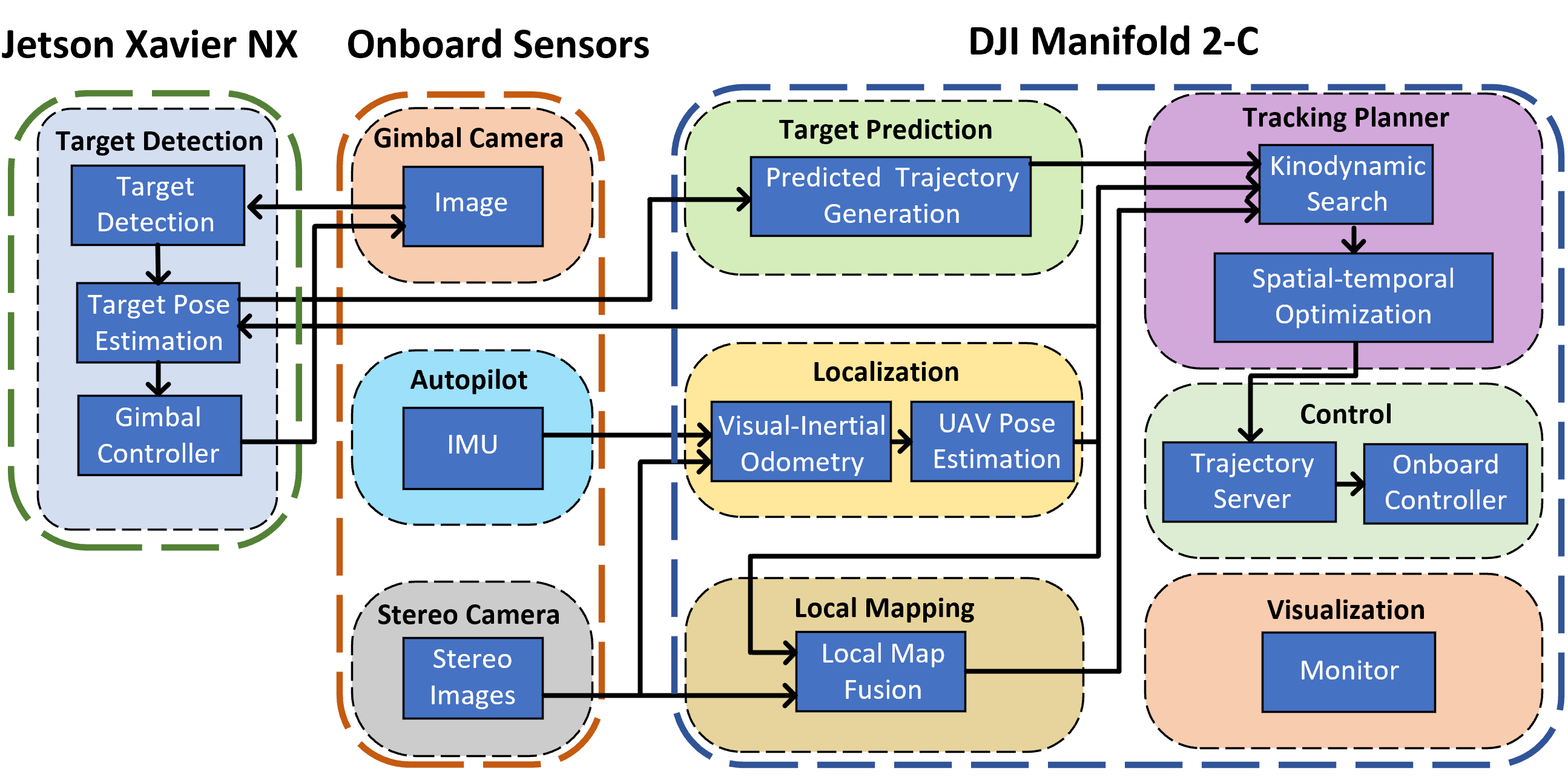}
		\captionsetup{font={small}}
		\caption{
			Software architecture of our quadrotor system. The sensing, planning and controlling modules run parallelly using onboard computing resources without external sensors.
		}
		\label{pic:system_architecture_full}
		\vspace{0.2cm}
	\end{figure*}
    
    On the whole, our systematic solution consists of five steps in a cascaded way: 1) the detection and location estimation of a human target (Sect.~\ref{sec:detection}) ; 2) the prediction of target future motion without its motion intention (Sect.\ref{sec:prediction}); 
	3) the tracking path searching (Sect.~\ref{sec:searching}) with an occlusion-aware mechanism; 4) the generation of safe and dynamically feasible tracking trajectories 
	(Sect.~\ref{sec:optimization}); 5) the construction of an onboard system (Sect.~\ref{sec:results}) with omnidirectional vision. 
    The proposed solution addresses most of the major challenges in aerial tracking. 
	
	The proposed methods are integrated into a quadrotor system (as shown in Fig.\ref{pic:system_architecture_full}). 
	Sufficient experiments in difficult real-world environments validate that the system can accurately detect and localize a human target, then aggressively and safely track it.  
	We also compare this work with Fast-Tracker~\cite{HZP2020tracker}. The results show that even the target detection and localization task becomes more challenging, the proposed system still achieves better tracking performance. 
	Contributions of this paper are:
	\begin{itemize}
		\item [1)] 
		A lightweight and intention-free human target detection and localization method based on deep learning and non-linear regression.
		\item [2)]
		Improvement of the integrated online tracking planning module in Fast-Tracker~\cite{HZP2020tracker} by incorporating an occlusion-aware mechanism.
		\item [3)]
		An onboard quadrotor system that integrates the proposed methods and the presentation of the systematic solution with extensive tests and evaluations.
	\end{itemize}

	\section{Related Work} 
	\label{sec:related_works}
	\subsection{Aerial Tracking Systems}
	 Some previous works \cite{JW2013noc, AG2014noc, HC2017noc} develop aerial tracking systems based on vision-based controllers that take the distance error from the quadrotor to the target obtained from the image space as the feedback. These methods fail to consider safety constraints, thus limiting the tracking scenarios only in wide-open areas.  
	 In recent years, several works\cite{TN2017elli, Bonatti2019JFR, BJ2020ICRA, JC2016tracking,HZP2020tracker} present more complete aerial tracking systems that consider obstacle avoidance, dynamical feasibility, and tracking efficiency. Nägeli et al.\cite{TN2017elli} rely on ground truth of the target's location and set a strong assumption on the obstacle's shape, limiting the application scenarios to artificial environments. Bonatti et al.\cite{Bonatti2019JFR} use a long range LiDAR sensor for mapping and use GPS for localization. However, bulky LiDAR sensors and GPS-based sensors can only be deployed on large outdoor platforms. Such system settings cannot work properly in indoor or cluttered environments like Fast-Tracker~\cite{HZP2020tracker} does. Jeon et al.\cite{BJ2020ICRA} require a prebuilt environmental map and assume that the target moves through a set of given via-points, making the system hard to be applied out of laboratories. Chen et al.\cite{JC2016tracking} and Fast-Tracker~\cite{HZP2020tracker} are able to handle unknown obstacle information and free moving target. Nevertheless, the two systems both depend on handcrafted marker-based target detection which oversimplifies the detection task. In short, the above aerial tracking systems are not applicable to challenging field applications.

	\subsection{Target Prediction and Tracking Trajectory Planning}
		 	
	The prediction module estimates the target's future motion as a guidance for trajectory planning. Similar schemes are proposed in \cite{AG2014noc, Bonatti2019JFR, TN2017elli} that adopt Kalman Filter with a specific motion model which is not accurate enough to precisely predict the future motion of a target with uncertain intention.
	Jeon et al. \cite{BJ2020ICRA} utilize covariant optimization that incorporates collision avoidance for target motion prediction. However, the collision term in the objective is computed from the Euclidean signed distance field (ESDF), consuming substantial computing resources. To estimate the target's future motion, Chen et al. \cite{JC2016tracking} adopt polynomial regression which is formulated as quadratic programming (QP). Fast-Tracker~\cite{HZP2020tracker} improves this method by adding dynamic constraints and time-variant confidence in the target predicted trajectory optimization.
	
	As for tracking trajectory planning, 
	Nägeli et al. \cite{TN2017elli} propose a receding horizon Model Predictive Control (MPC) planner that considers visibility to targets, collision avoidance, and occlusion minimization. Similarly, the planning module in \cite{Bonatti2019JFR} optimizes a set of cost functions based on covariant gradient descent. However, the optimization formulations both contain multiple non-convex terms, and the solution might fall into a local minimum.
	Jeon et al. \cite{BJ2020ICRA} propose a bi-level approach composed of a graph-search based path planner and a hard-constrained trajectory optimizer. Because a dense local graph has to be built and traversed in every search process, the graph-search method is highly time-consuming. Chen et al. \cite{JC2016tracking} optimize the tracking error and high-order derivatives of the trajectory within a flight corridor. Nevertheless, this work fails to account for occlusion against obstacles. 
	
	\subsection{Human Detection and Localization}
	Several previous works\cite{Bonatti2019JFR, HC2017noc, Huang2018Openpose} focus on detecting 2D targets utilizing monocular images and recovering their 3D location. Bonatti et al.\cite{Bonatti2019JFR} combine MobileNet\cite{MobileNet} and Faster-RCNN\cite{ren2015faster} for target detection. A ray-casting module is then proposed to project the 2D bounding box onto a height map provided by an onboard LiDAR sensor. As stated above, the LiDAR sensor is a heavy load in target tracking scenarios. Cheng et al.\cite{HC2017noc} set a tracking scenario that the quadrotor flies at a high altitude and subsequently considers the human target as a point. The quadrotor's altitude is then approximated as the target's depth in the images and utilized for 3D location estimation. However, the high-altitude flight prevents the method from applying to indoor or narrow environments. Huang et al. \cite{Huang2018Openpose} use a sequence-to-sequence network to estimate 3D pose from a sequence of 2D joints obtained from skeleton detection and then define an optimization function to estimate the target's location. 
	The above methods are mainly used in aerial cinematography, where target pose estimation is required for planning the shooting angle.
	As for aerial tracking, there is no pressing need for target pose estimation. Compared with \cite{Huang2018Openpose}, although we use skeleton detection method as well, we directly estimate the location of the target without pose estimation to save computing resources.
	
	To enable active vision and improve noisy target observations, previous works\cite{Bonatti2019JFR, HC2017noc, Galvane2018, TN2017elli, Huang2018Openpose, JOUBERT2016} equip an onboard 3-DOF gimbal camera. The target can be kept at the desired position within the camera plane by adopting closed-loop controllers or optimization methods. 
	In fact, due to the fact that target does not move drastically in the vertical direction in most tracking scenarios, controlling its horizontal position in the camera plane is sufficient to avoid target loss.
	We thus customize a lightweight 1-DOF gimbal instead, ensuring the agility of the proposed quadrotor system. We also equip an electrical slip ring structure to enable free rotation of the gimbal to enhance target perception capability.

	\begin{figure}[t]
	\centering
	\includegraphics[width=1\linewidth]{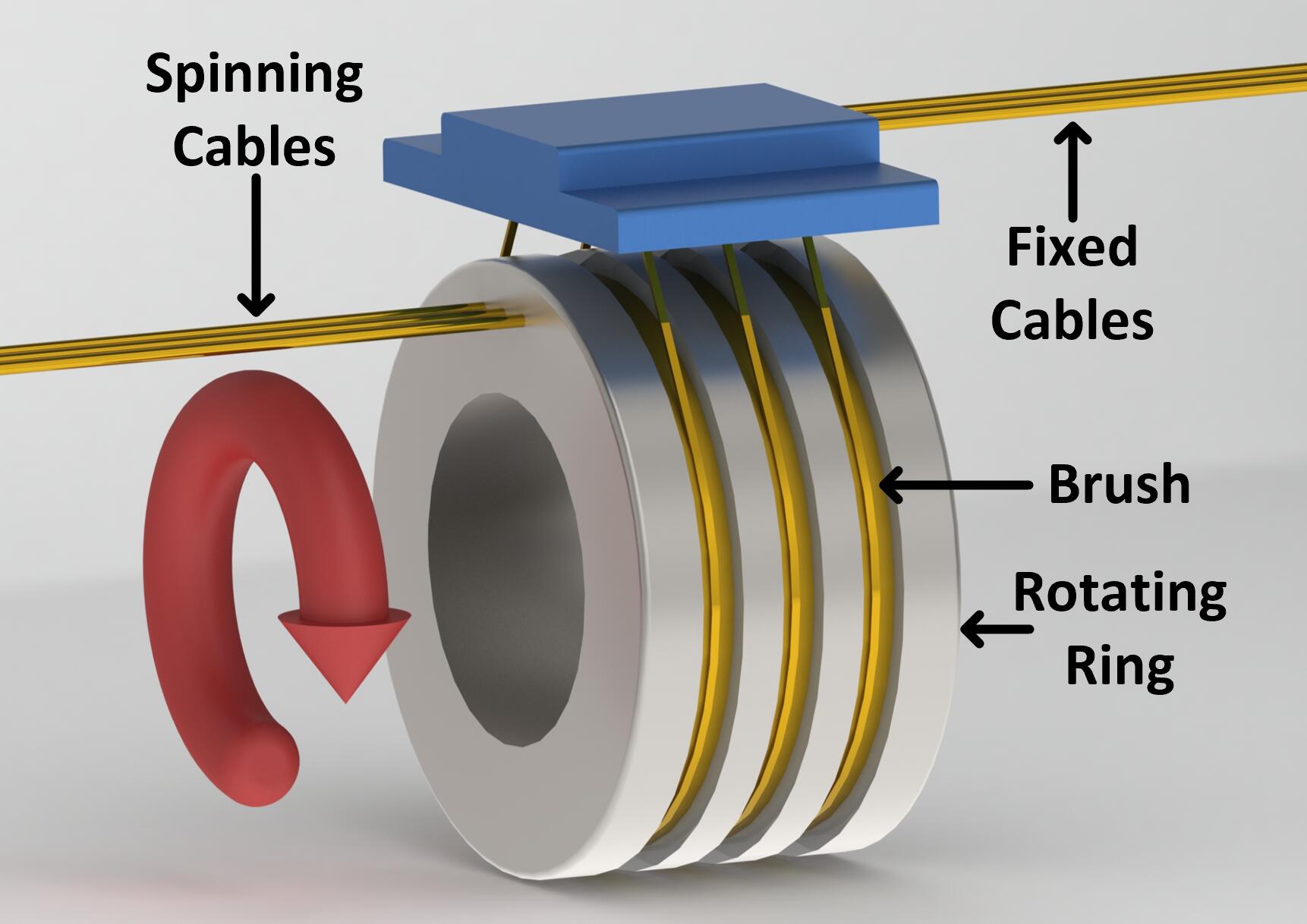}
	\captionsetup{font={small}}
	\caption{
		The schematic diagram of the slip ring.
	}
	\label{pic:slip_ring}
	\vspace{-0.3cm}
	\end{figure}		

	\section{Overview}
	\label{sec:formulation}
	\subsection{Problem Formulation}
	The overall task is to control a quadrotor to detect, localize, and track a human target with uncertain intention in unknown environments.
	Let $\hat{\mathbf{p}}_t^w = \{x_{t}^w, y_{t}^w, z_{t}^w\}\in \mathbb{R}^3$ be the the estimated 3D position of the target in the world frame (Sect.~\ref{sec:detection}), and $\hat{P}_{t}(t) : (t_c, t_p]  \rightarrow \mathbb{R}^3 $ be the target predicted trajectory from the current time $t_c$ to $t_p$  (Sect.~\ref{sec:prediction}). For the quadrotor planning module, we use voxel occupancy grids denoted as ${\mathcal{G} : \mathbb{R}^3 \rightarrow [0, 1]}$ to represent the environment. Using $\hat{P}_{t}$ as a guidance, the front-end path searching method generates a grid path with kinodynamic information (Sect.~\ref{sec:searching}). Utilizing the end-derivatives and the flight corridor $\mathcal{F}$ obtained from the grid path, the back-end optimizer generates a piecewise-polynomial trajectory ${P}_{q}(t) \in \mathbb{R}^3$ for the quadrotor to track the target (Sect.~\ref{sec:optimization}).

	For the tracking scenarios, we make a set of simplifying assumptions: 
	\begin{itemize}
	\item [1)] 
	The target's motion in the vertical direction is negligible during the tracking process. 
	\item [2)]
	The upper body length of the target remains roughly the same.
	\item [3)] 
	The target moves smoothly with bounded velocity and acceleration.
	\end{itemize}

	\subsection{Active Vision Design}
	According to assumption 1, we focus on controlling the target's horizontal position in the camera plane. An 1-DOF gimbal is thus deployed to control the camera's yaw angle so that the target is positioned in the middle of the frame in the horizontal direction. We use a PI controller to control the camera independently of the quadrotor's motion planning. 
	
	Since the camera and the gimbal are fixed, the camera's connection cable will limit its rotation. To enable the gimbal motor's free rotation, we replace the connection cable with a customized electrical slip ring. A slip ring's primary components are a rotating ring and stationary brushes that rub with the ring, as illustrated in Fig.\ref{pic:slip_ring}. Data transmission between the spinning and stationary cables is realized for they are connected with the rotating ring and the brushes, respectively. Therefore, the camera can rotate freely while remaining connected with the onboard computer.

	Credit to the active vision, we now keep the quadrotor facing the direction of its linear velocity rather than facing the target. As a result, the target and the surrounding obstacles can be well observed at the same time.

	\begin{figure}[t]
	\centering
	\includegraphics[width=1\linewidth]{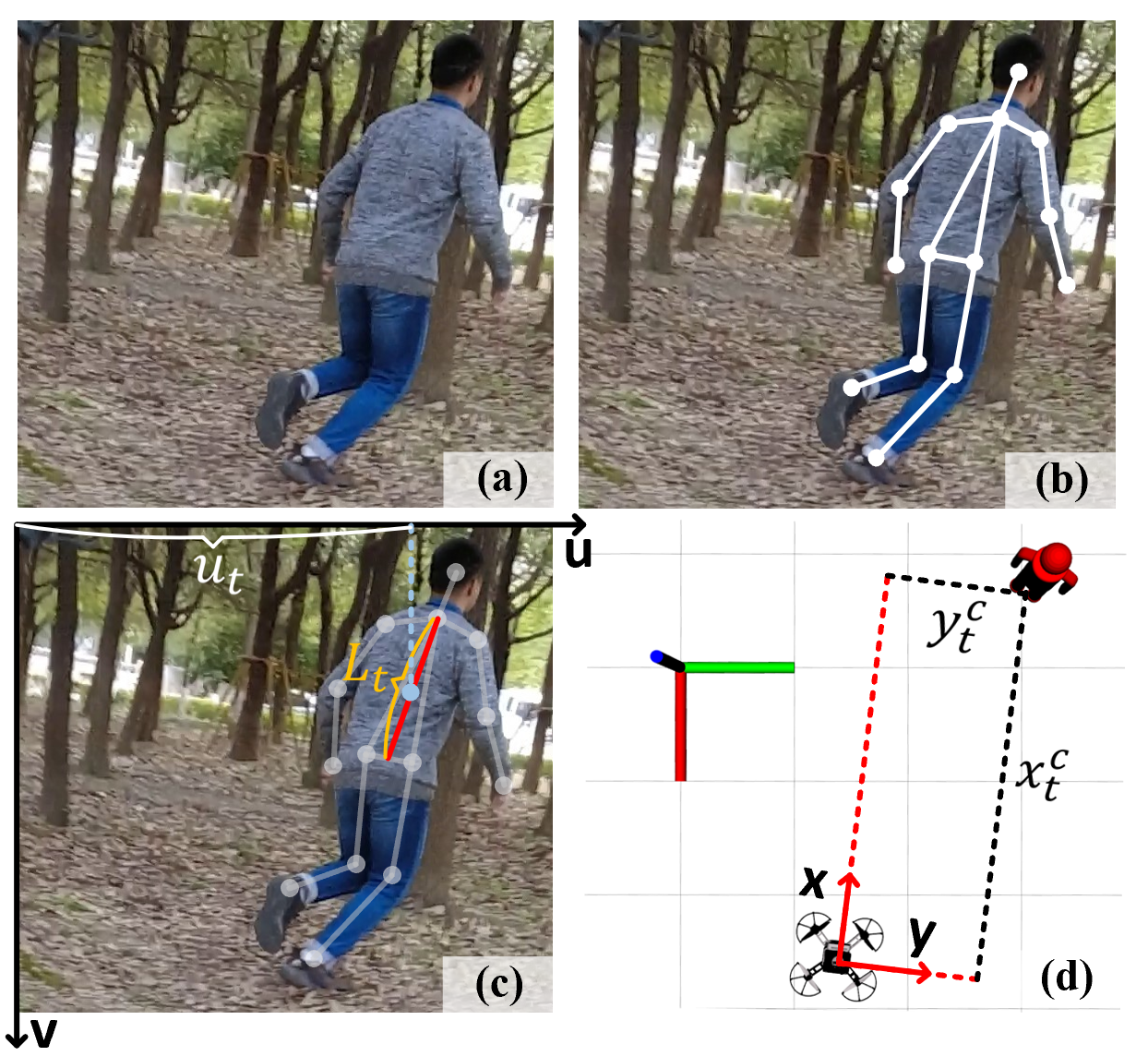}
	\captionsetup{font={small}}
	\caption{
		a) A raw monocular image for target detection. b) The target detection result. c) The extracted features ${L}_t$ and $u_t$ in the camera plane. d). The 3D location estimation result.
	}
	\label{pic:detection}
	\vspace{-0.45cm}
	\end{figure}	
	
	\section{Human Detection with Location Regression}
	This section elaborates on the human target detection and localization methods. Utilizing monocular images obtained from the gimbal camera, this module outputs $\hat{\mathbf{p}}_t^w= \{x_{t}^w, y_{t}^w, z_{t}^w\}$. The entire module is shown in Fig.\ref{pic:detection}.
    \label{sec:detection}
    \subsection{Human Target Detection}
    For human detection, we apply OpenPose\cite{Openpose}, a deep learning based 2D pose estimation method which detects 2D joints, namely the head, shoulders, knees, etc.. OpenPose adopts a bottom-up parsing architecture that jointly learn part locations and their association via Part Affinity Fields(PAFs) and presents remarkable performance and efficiency. We equip OpenPose with TensorRT\footnote{https://github.com/NVIDIA/TensorRT} inference optimizer to ensure real-time capability on the onboard platform. 

	Due to assumption 2, the target's upper body length can be utilized as an invariant feature. We take the the distance from the hip to the neck in the camera plane as the upper body length and utilizes it for target localization.
	
	\subsection{Non-linear Regression Based Target Localization}
	Let ${L}_t$ be the detected upper body length in the camera plane. We also utilize the $u$ coordinate value of the target's center in the camera plane denoted as $u_t$ for localization. Non-linear functions are used to establish the mapping from ${L}_t$ and ${u}_t$ to the estimated target location in the camera frame denoted as $\hat{\mathbf{p}}_t^c= \{x_{t}^c, y_{t}^c, z_{t}^c\}$. Especially, $z_{t}^c$ is set to be a constant because of assumption 1. We then define the following functions with a set of undetermined parameters:
	\begin{equation}
	\left\{
	\begin{aligned}
	& x_{t}^c = \lambda_{1}e^{k_1{L}_t} + \lambda_{2}e^{k_2{L}_t}, \\
	& y_{t}^c = (\lambda_{3}e^{k_3{u}_t} + \lambda_{4}e^{k_4{u}_t})(ax_{t}^c + b).	\\
	\end{aligned}
	\right.
	\end{equation}	
	In practice, we make a dataset under the ground truth obtained from a motion capture system (mocap) to train these parameters. $\hat{\mathbf{p}}_t^c$ is converted to $\hat{\mathbf{p}}_t^w$ through coordinate transformation in the end. 
	
\begin{figure}[t]
	\centering
	\includegraphics[width=1\linewidth]{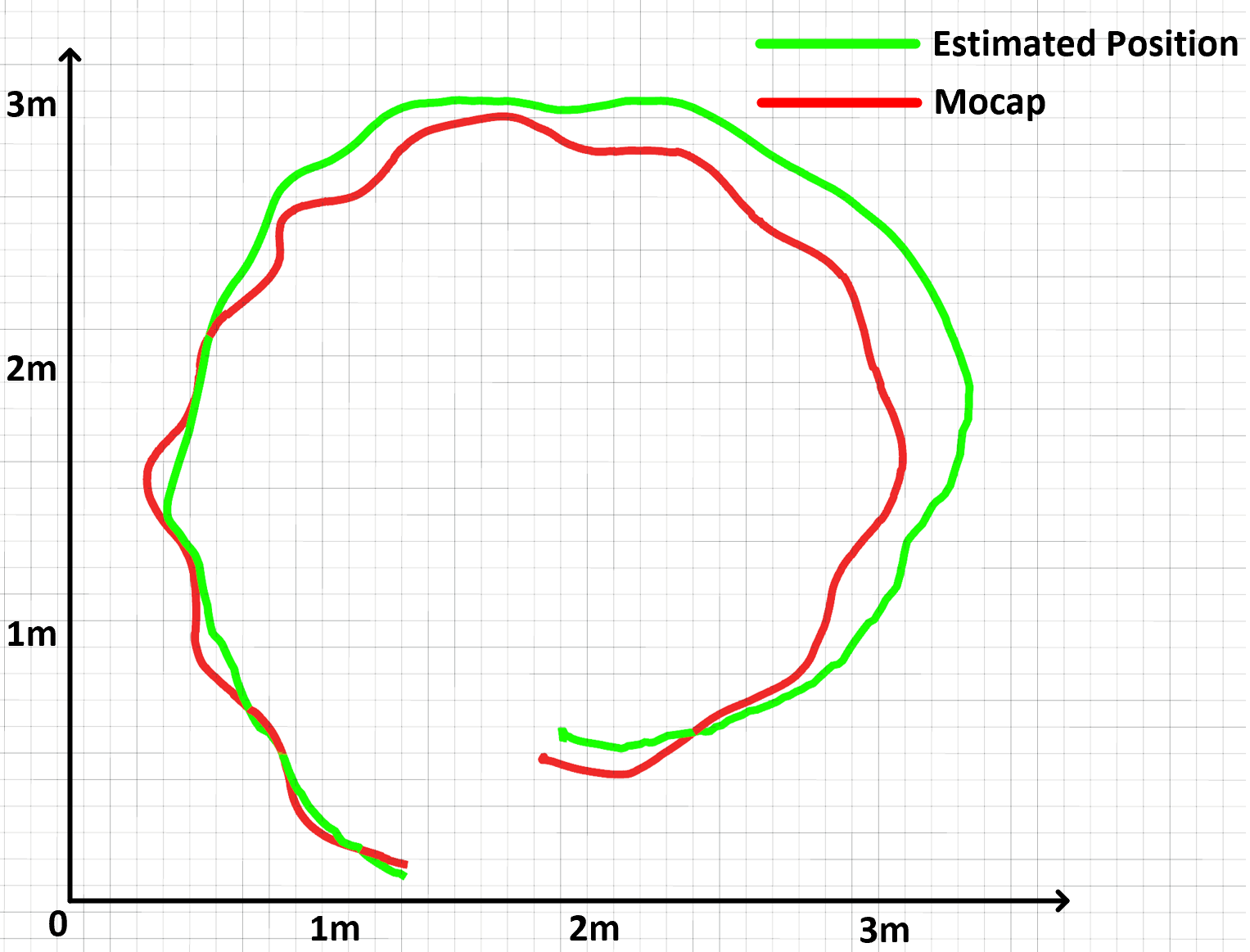}
	\captionsetup{font={small}}
	\caption{
		Tests of the target Localization method under a mocap.
	}
	\label{pic:motion_capture}
	\vspace{-0.7cm}
	\end{figure}

	We test the performance of the location regression method, as shown in Fig.\ref{pic:motion_capture}. The average distance error is about 20cm, which is acceptable for the tracking scenarios. The results (Sect.~\ref{sec:results}) also confirm that this method is practical and effective. 

	\begin{figure*}[t]
	\centering
	\begin{subfigure}{0.245\linewidth}
		\includegraphics[width=1\linewidth]{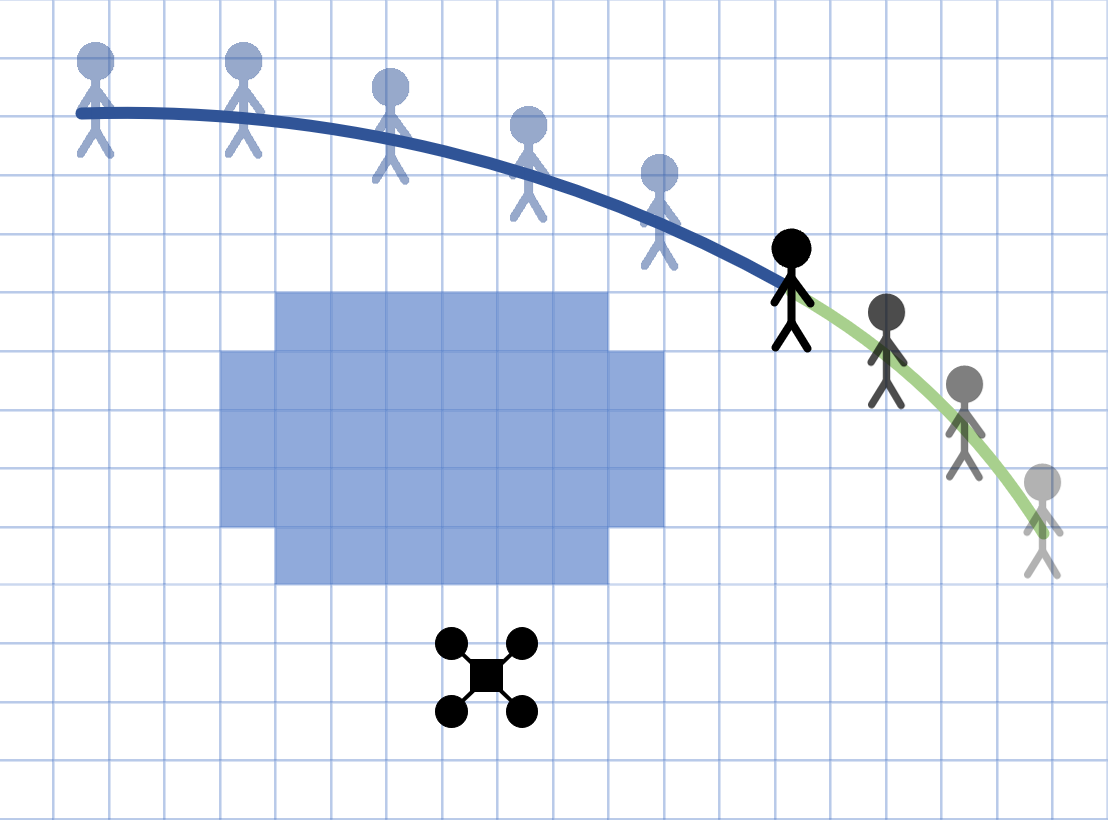}
		\captionsetup{font={small}}
		\caption{Target motion prediction.}
		\label{pic:planning_a}
	\end{subfigure}
	\begin{subfigure}{0.245\linewidth}
		\includegraphics[width=1\linewidth]{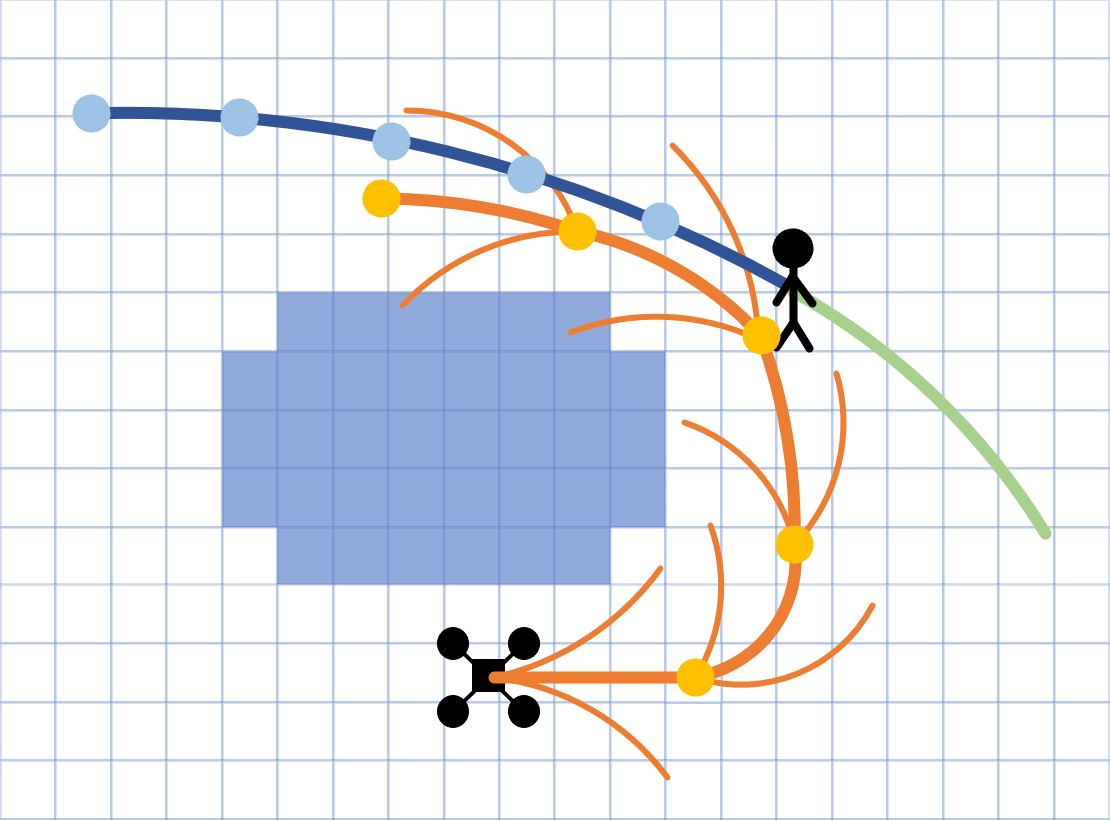}
		\captionsetup{font={small}}
		\caption{Kinodynamic path searching.}
		\label{pic:planning_b}
	\end{subfigure}
	\begin{subfigure}{0.245\linewidth}
		\includegraphics[width=1\linewidth]{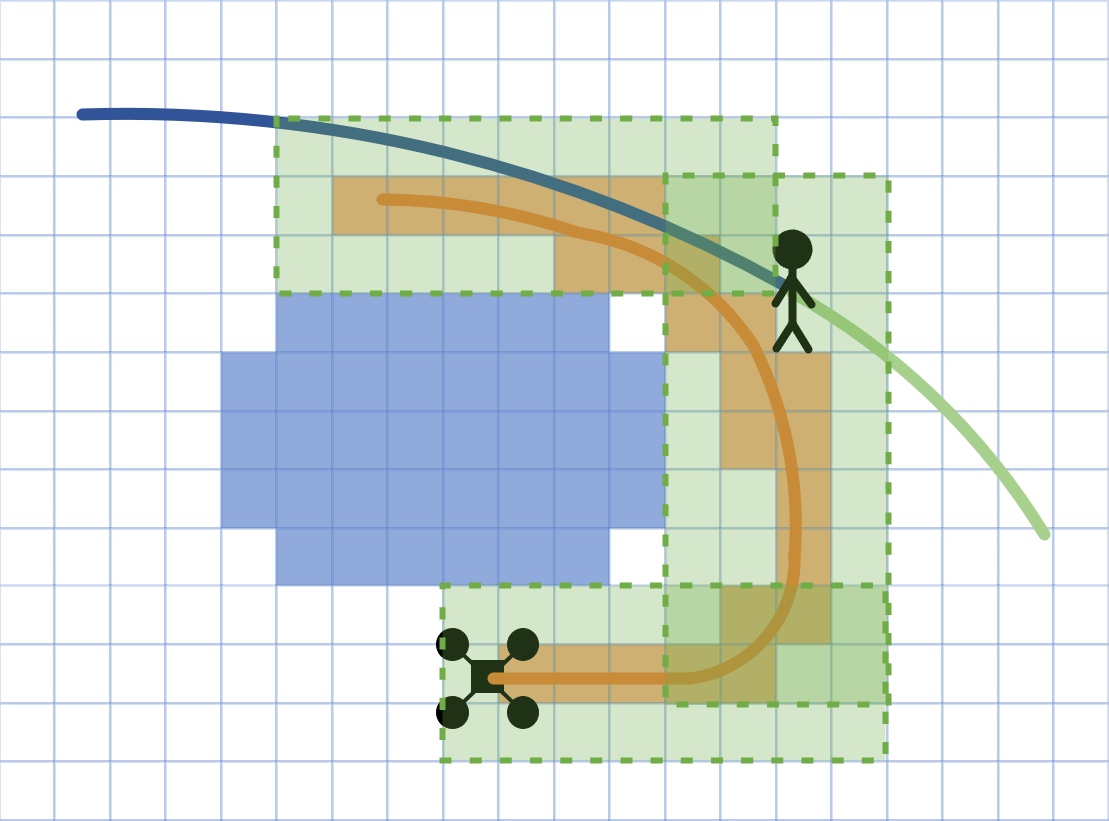}
		\captionsetup{font={small}}
		\caption{Flight corridor generation.}
		\label{pic:planning_c}	
	\end{subfigure}
	\begin{subfigure}{0.245\linewidth}
		\includegraphics[width=1\linewidth]{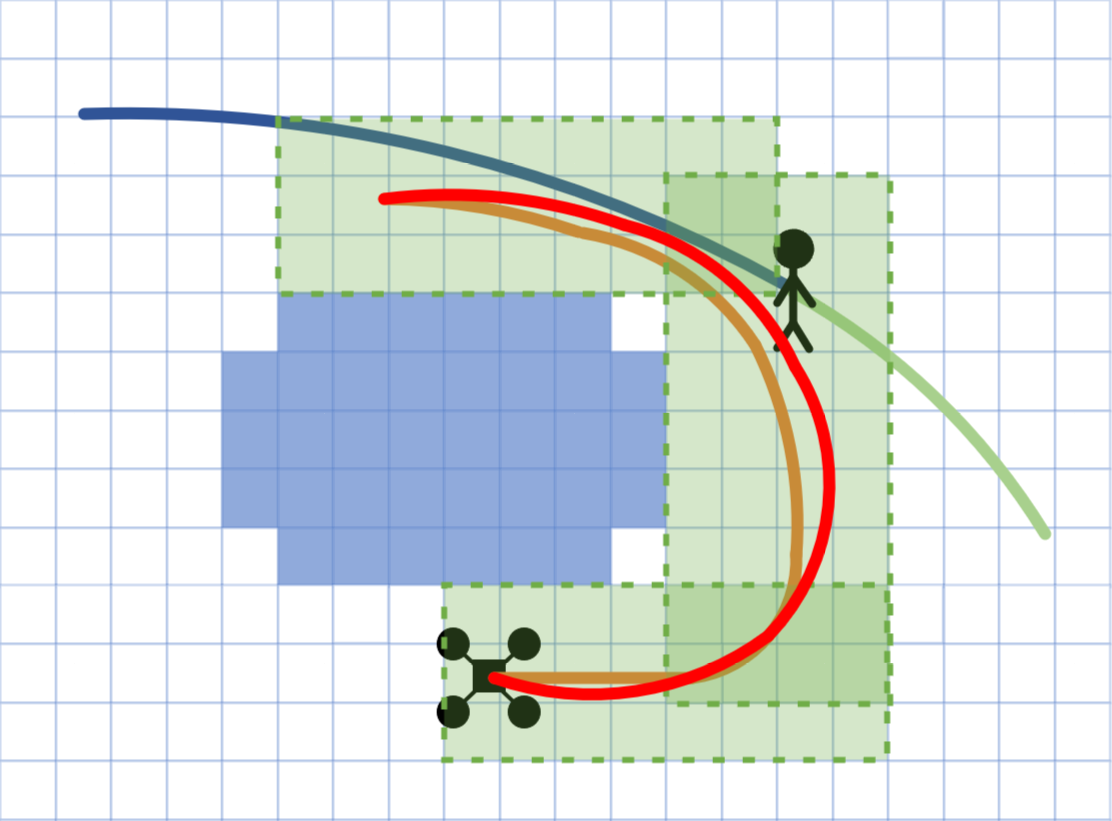}
		\captionsetup{font={small}}
		\caption{Trajectory optimization.}
		\label{pic:planning_d}	
	\end{subfigure}
	\captionsetup{font={small}}
	\caption{a) A target predicted trajectory $\hat{P}_{t}$ (the blue curve) is generated based on past location estimations $\hat{\mathbf{p}}_t^w$ (the black markers).
		b) A kinodynamic grid path consisted of motion primitives (the orange curves) is generated for the quadrotor to track the target.
		c) A flight corridor $\mathcal{F}$ (the sequence of green cubes) is generated based on the grid path.
		d) A spatial-temporal trajectory ${P}_{q}$ (the red curve) is generated within the flight corridor.
	}
	\vspace{0cm}
	\label{pic:planning}
	\end{figure*}

	\section{Online Tracking Planning}
	\label{sec:planning}
	In this section, we introduce the tracking planning module of Fast-Tracker~\cite{HZP2020tracker} shown in Fig.\ref{pic:planning} and the proposed improvements on it.
	\subsection{Target Motion Prediction}
	\label{sec:prediction}
	This method updates the target predicted trajectory $\hat{P}_{t}(t) : (t_c, t_p]$ utilizing the past target estimations acquired from Sect.~\ref{sec:detection}. Polynomial regression is adopted to fit the target motion estimations. We specially use Bernstein basis polynomial (B\' ezier curve) to enforce dynamical constraints due to its convex hull property and hodograph property (described in \cite{gao2018online}). The target predicted trajectory is presented as: 
	\begin{equation}
	\hat{P}_{t}(t)=\sum_{i=0}^nc_ib_n^i(t),
	\end{equation}
	where each $b_n^i(t) = \left(\begin{aligned}&n\\&i\end{aligned}\right)\cdot t^{i} \cdot(1-t)^{n-i}$ is an n-degree Bernstein polynomial basis, and $[c_0, c_1, ..., c_n]$ is the set of control points of the B\' ezier curve. 
	
	 The object function $J_{pre}$ is formulated as QP that includes the residual term $J_{d}$ and regularization term $J_{r}$:
	\begin{equation}
	J_{pre} = J_{d} + J_{r},
	\end{equation}	

	where $J_{d}$ minimizes the distance residual between $\hat{P}_{t}$ and the past target location estimations. $J_{r}$ is an acceleration regulator for avoiding over-fitting. 

	Based on assumption 3, the predicted velocity and acceleration are constrained in pre-defined ranges to ensure the dynamical feasibility. Time-variant weights are also added to decrease the confidence of old target location estimations. The obtained trajectory is then extrapolated to $(t_c, t_p]$ for target motion prediction. The detailed formulations are described in \cite{HZP2020tracker}. 

	A target-relocating strategy is proposed in Fast-Tracker~\cite{HZP2020tracker} to make the quadrotor actively explore and rediscover the target after losing it. With the active vision, we keep the gimbal rotating to search for the target during the relocating process. This mechanism improves the target-relocating strategy's effectiveness and is presented in the results (Sect.~\ref{sec:results}).
	
	\subsection{Occlusion-aware Tracking Path Searching}
	\label{sec:searching}	
	The path searching method is originated from hybrid A* algorithm\cite{DD2008HybridAstar}. Let the position and velocity vector $\mathbf{x}={(p_{x},p_{y},p_{z},v_{x},v_{y},v_z)}^{T}$ be the state of the quadrotor. The quadrotor's acceleration is taken as the control input $\mathbf{u}$ and discretized to generate motion primitives, namely the nodes in hybrid A* algorithm.

	The cost function of each node is written as $f_c=g_c+h_c$, where $g_c$ represents the actual cost from the initial state $\mathbf{x_0}$ to the current expanded state  $\mathbf{x_c}$ and $h_c$ means the heuristic cost. 
	$g_c$ balances the control-effort $\mathbf{u}$ and the time $T$ of a trajectory by minimizing a energy-time cost defined as: 	
	\begin{equation}
	J_{t} =\int_{0}^{T}{||\mathbf{u}(\tau)||}^2d\tau+\mathbf{\rho}T,
	\end{equation}	
	where $\mathbf{\rho}$ is the weight term. In practice, this cost function is discretized to be the sum of the expanded nodes' cost.
	
	$h_c$ considers the heuristic cost from $\mathbf{x_c}$ to the goal state $\mathbf{x_g}$. We denote the dynamic cost from $\mathbf{x_c}$ to $\mathbf{x_g}$ as $D(\mathbf{x_c},\mathbf{x_g})$. It is obtained from solving an optimal boundary value problem (OBVP) proposed in \cite{zhou2019robust}. To make the path search both reliable and forward-looking, we use the weighted value of the target's current state $\mathbf{x_{tc}}$ and predicted state $\mathbf{x_{tp}}$ obtained from $\hat{P}_{t}$ as the goal state $\mathbf{x_g}$.  
	Moreover, a time penalty $P_{time}$ is added to $h_c$ speed up searching.

\begin{figure}[t]
	\centering
	\includegraphics[width=1\linewidth]{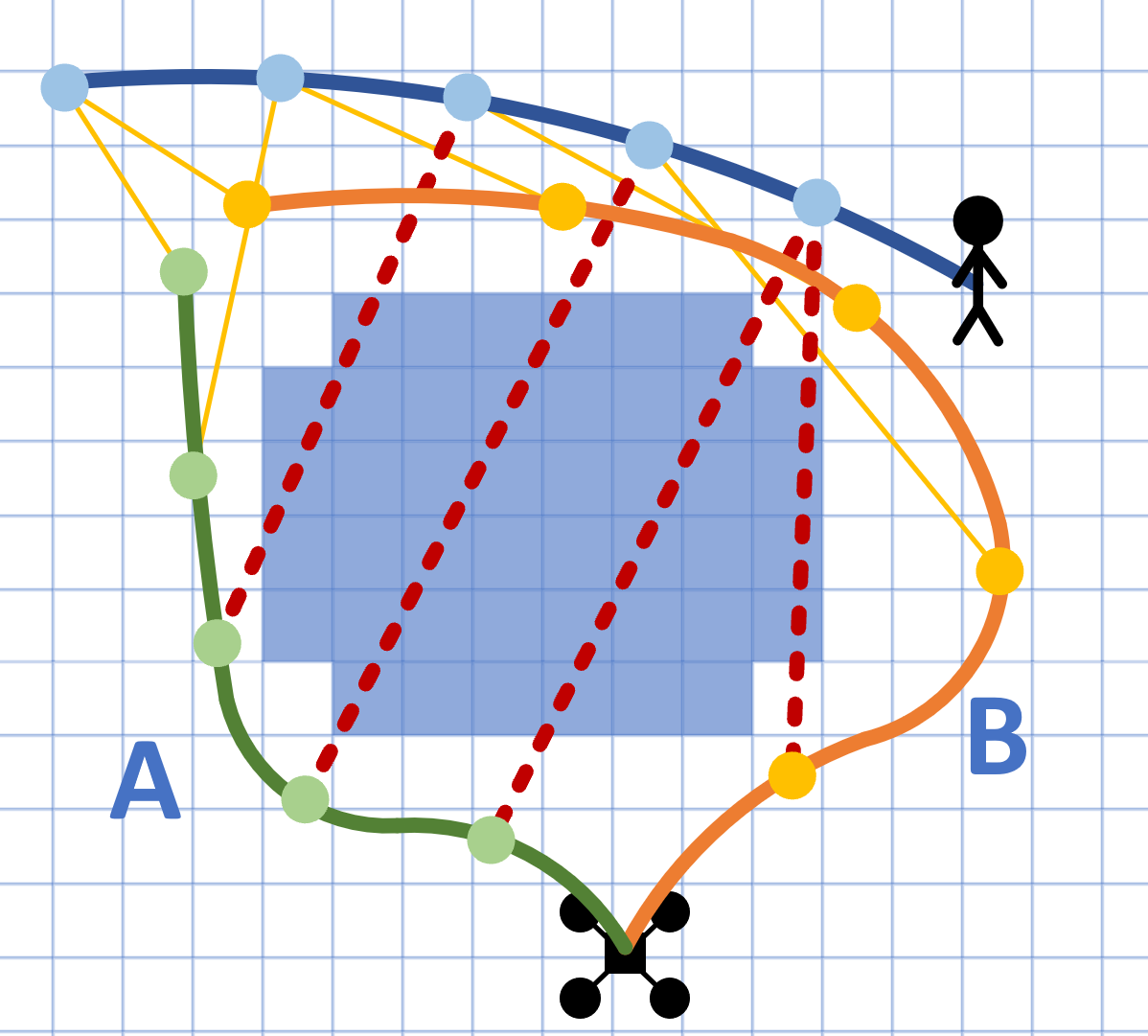}
	\captionsetup{font={small}}
	\caption{
		Illustration of the occlusion-aware mechanism. Two topology paths (the green and orange curves) are generated based on the target predicted trajectory (the blue curve). Although topology A is shorter than B, it is suboptimal because it will cause more occlusion than B in the future tracking process. Credit to the occlusion penalty, topology B will be chosen.
	}
	\label{pic:occlusion_aware}
	\vspace{-1.2cm}
\end{figure}

	In this paper, occlusion check is conducted between $\mathbf{x_c}$ and $\mathbf{x_{tp}}$ in each expanding process. We claim that occlusion will happen if the line from $\mathbf{x_c}$ to $\mathbf{x_{tp}}$ collides with obstacles, and this node cannot pass the occlusion check. An additional term $P_{occ}$ is then added to $h_c$ to penalize such nodes. This mechanism is illustrated in Fig.\ref{pic:occlusion_aware}. In general, $h_c$ is formulated as follows:
	\begin{equation}
	h_c = D(\mathbf{x_c},\mathbf{x_g})+P_{time}+P_{occ},
	\end{equation}
	
	\begin{figure}[t]
	\centering
	\includegraphics[width=1\linewidth]{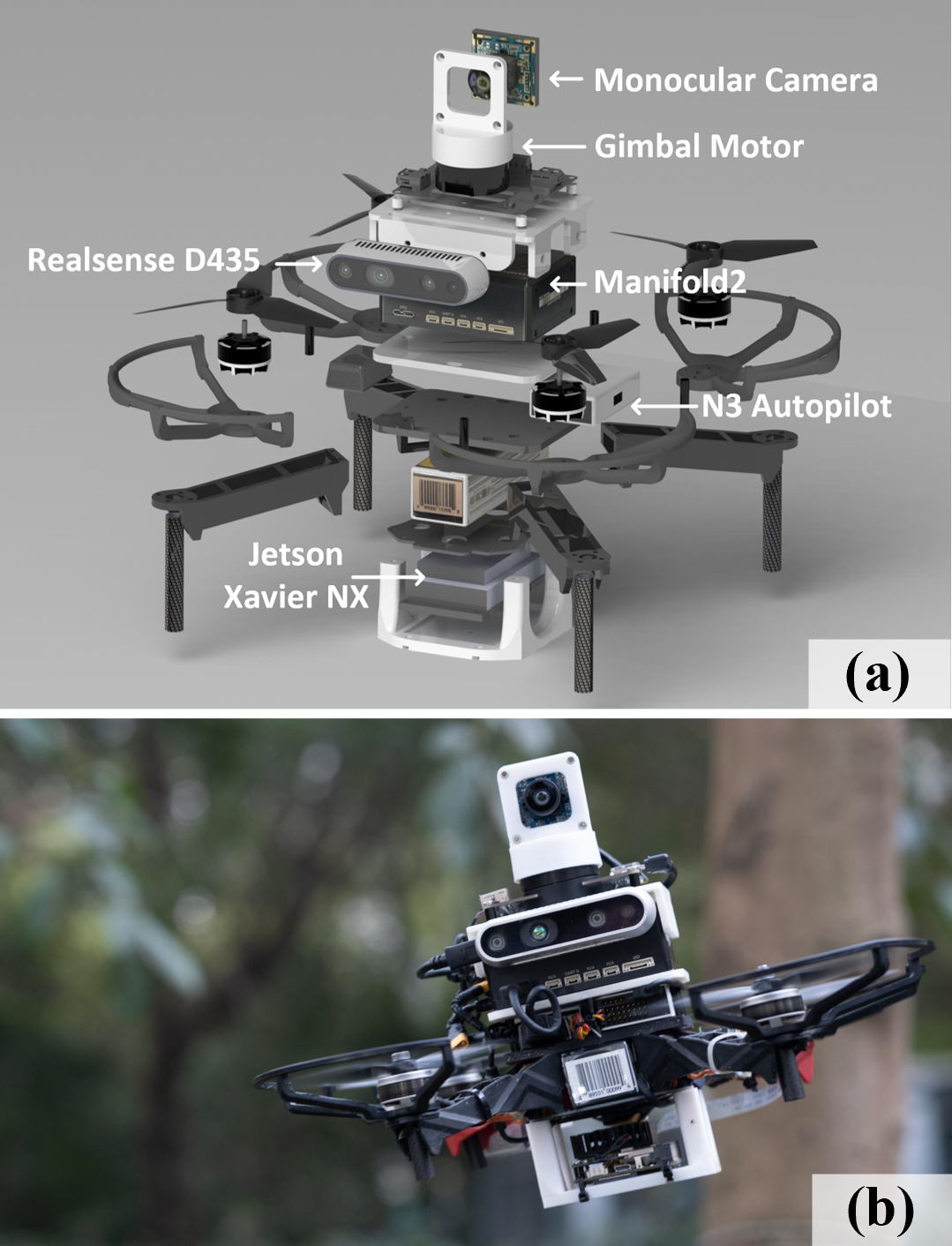}
	\captionsetup{font={small}}
	\caption{
		a) Exploded diagram. b) Real-world quadrotor system.
	}
	\label{pic:quadrotor}
	\vspace{0.2cm}
	\end{figure}

	\begin{figure*}[t]
	\centering
	\includegraphics[width=1\linewidth]{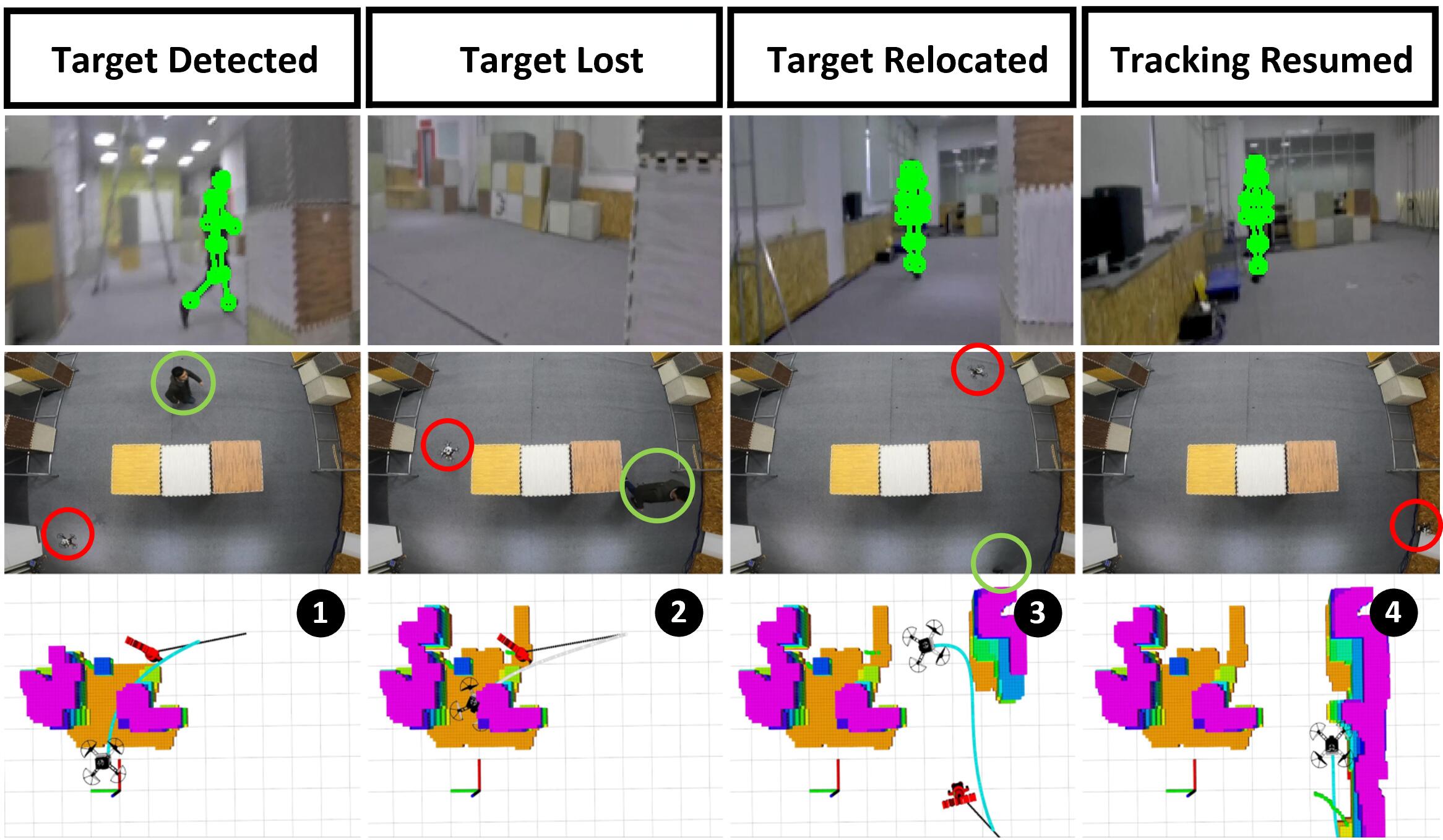}
	\captionsetup{font={small}}
	\caption{
		The schematic diagram of the target relocation strategy. In the bottom visualization diagrams, the black curves represent the target predicted trajectory, and the cyan curves represent the quadrotor's tracking trajectory. 1) The quadrotor tracks the target based on target location estimation and motion prediction. 2) After the obstacles occlude the target, the quadrotor flies along the re-locating trajectory (grey curve). Meanwhile, the gimbal camera keeps rotating to search for the target 3) The target is re-located by the gimbal camera. 4) A new trajectory is generated for the quadrotor to carry on tracking the target. 
	}
	\label{pic:relocate_strategy}
	\vspace{0.1cm}
    \end{figure*}


	\subsection{Safe Tracking Trajectory Generation}	
	\label{sec:optimization}
	We use the efficient spatial-temporal piece-wise polynomial trajectory generator\cite{wang2020DDP} for back-end optimization. This method derives an analytic gradient for the waypoints $q_w$ and piece times $T$ of with linear complexity. Leveraging the exact gradient, the decision variables are reduced into $q_w$ and $T$.  

	Chiefly, a flight corridor $\mathcal{F}$ is generated based the grid path obtained from the path searching method, defined as:
	\begin{equation}
	\mathcal{F} = \bigcup\limits_{i=1}^M \mathcal{C}_i,
	\end{equation} 
	where each $\mathcal{C}_i$ is a finite cube:
	\begin{equation}
	\mathcal{C}_i = \{ x \in \mathbb{R}^3  | \mathbf{A}_{c_{i}} x \leq b_{c_{i}} \}.
	\end{equation}
	For such a $\mathcal{F}$, the start and goal position is located in $\mathcal{C}_1$ and $\mathcal{C}_M$, respectively. This method assign each intermediate waypoint $q_{w_i}$ in the intersection $\mathcal{C}_{i}  \cap \mathcal{C}_{i+1}$ to ensure safety.
	
	The optimization formulation trades off between the smoothness $J_S$, the corridor constraint $J_F$, and the aggressiveness penalty $J_D$:
	\begin{equation}
	J_{\sum}(q,T)={J_S}(q,T)+J_F(q)+J_D(q,T).
	\end{equation}
	$J_S$ minimizes the magnitude of jerk along the piecewise trajectory in 3 dimensions to improve smoothness. $J_F$ is a logarithmic barrier term to ensure that each $q_{w_i}$ is constrained in $\mathcal{C}_{i} \cap \mathcal{C}_{i+1}$, defined as: 
	\begin{equation}
	J_F(q_w) = -\kappa \sum_{i=1}^{M-1}\sum_{j=i}^{i+1}\mathbf{1}^T \ln \left[ b_{c_{j}} -\mathbf{A}_{c_{j}} q_{w_i}  \right].
	\end{equation}
	where $\kappa$ is a constant coefficient, $\mathbf{1}$ an all-ones vector and $\ln \left[ \cdot \right]$ the entry-wise natural logarithm.
	
	The last term $J_D$ adjusts the aggressiveness of the whole trajectory. It is defined as:
	\begin{align}
	&J_D(q_w,T)=\rho_t\sum_{i=1}^{M}T_i+           \\
	&\rho_v\sum_{i=1}^{M-1}  l( {\left\| \frac{q_{w_{i+1}}-q_{w_{i-1}}}{T_{i+1}+T_{i}} \right\|}^2 - {v_m}^2)+ \nonumber \\
	&\rho_a\sum_{i=1}^{M-1}l({\left\|  \frac{(q_{w_{i+1}}-q_{w_i})/T_{i+1}-(q_{w_i}-q_{w_{i-1}})/T_i}{(T_{i+1}+T_i)/2}   \right\|}^2 -a_m^2) \nonumber,
	\end{align}
	where $l(x) = max(x,0)^3$, $v_m$ is the maximum velocity and $a_m$ is the maximum acceleration.	$\mathbf{\rho}_t$ prevents the entire duration from growing too
	large. $\mathbf{\rho}_v$ and $\mathbf{\rho}_a$ limit the velocity and acceleration to prevent the tracking trajectory from being too aggressive.
	
	\section{Results}
	\label{sec:results}
	\subsection{Implementation Details}
	The quadrotor system built for real-world experiments is shown in Fig.\ref{pic:quadrotor}. It is equipped with Intel RealSense
	D435 depth camera\footnote{https://www.intelrealsense.com/depth-camera-d435/} for self localization\cite{qin2017vins} and mapping. 
	The localization, mapping, and motion planning algorithms run on Manifold2\footnote{https://www.dji.com/manifold-2} with i7-8550U CPU, while the target detection algorithm runs on Jetson Xavier NX\footnote{https://developer.nvidia.com/embedded/jetson-xavier-nx} with GPU performance of 21 TOPS. The horizontal gimbal camera is driven by a brushless motor. The quadrotor is $300\times300mm$ in size and only weighs $1.4kg$ including a 4s battery, which is dexterous enough for most tracking scenarios.
    The average computing times are $40$ms and $20$ms, respectively for the target detection and motion planning modules. The re-planning frequency of the whole pipeline is set to be $13$Hz.

	\subsection{Real-world Experiments}
	This section presents extensive experiments in several unknown cluttered environments to test the proposed approach's robustness and tracking performance. 
	
\begin{figure}[t]
	\centering
	\begin{subfigure}{1\linewidth}
		\centering
		\includegraphics[width=1\linewidth]{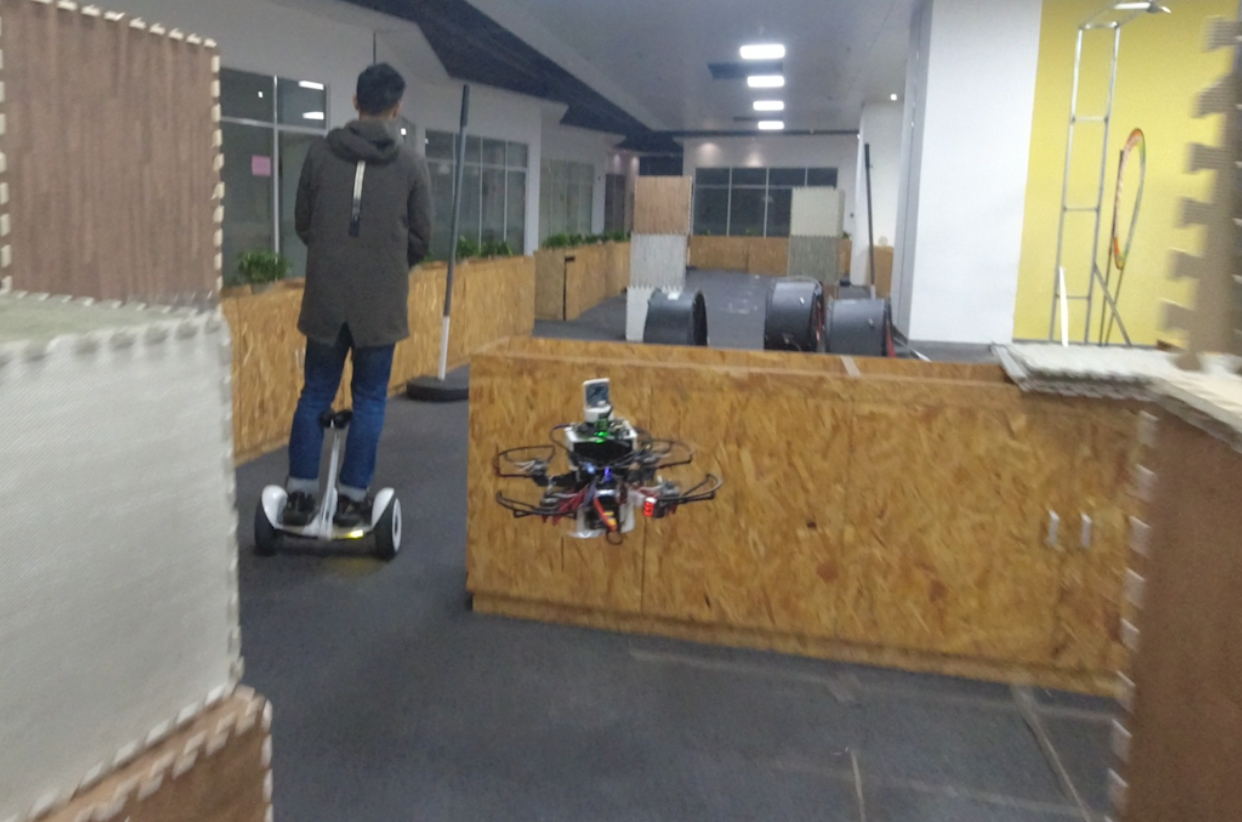}
		\captionsetup{font={small}}
		\caption{The indoor experiment in a narrow and cluttered corridor.}
		\label{pic:indoor_exp1}
	\end{subfigure}
	\begin{subfigure}{1\linewidth}
		\centering
		\includegraphics[width=1\linewidth]{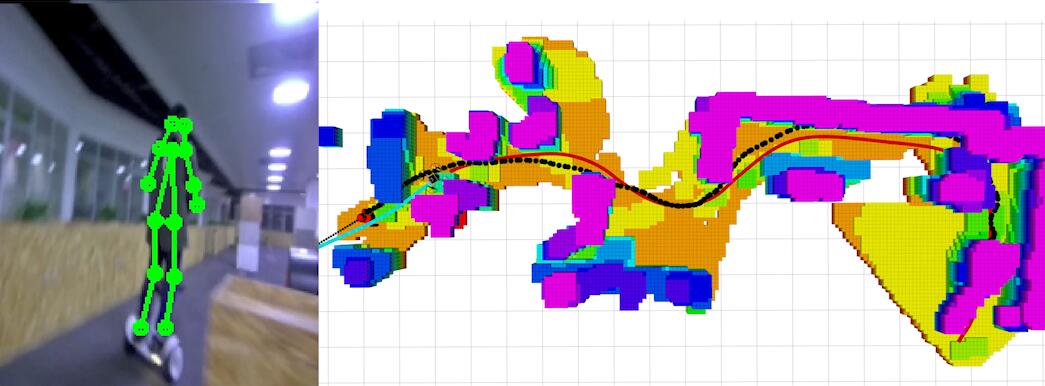}
		\captionsetup{font={small}}
		\caption{The third perspective and the visualization of the indoor experiment. }			
		\label{pic:indoor_exp2}
	\end{subfigure}
	\captionsetup{font={small}}
	\caption{
		Demonstration of the indoor experiment. 
	}
	\vspace{-0.4cm}
	\label{pic:indoor_exp}
	\end{figure}

	The first experiment is carried out in a narrow and cluttered corridor, as shown in Fig.\ref{pic:indoor_exp1}. The target drives a balance car to move agilely in the corridor. The cramped environment requires the quadrotor to respond quickly and re-plan the tracking trajectory in case of an emergency.  As a result, the quadrotor flies safely in this narrow environment and closely follows the target.

	Another experiment demonstrates the improved target-relocating strategy, as shown in Fig.\ref{pic:relocate_strategy}. 
	In this experiment, the quadrotor loses the target due to the its sudden turn. Moreover, the target turns again and moves fast in another direction, making it hard to be rediscovered. As a result, the target is still relocated because the gimbal camera keeps searching for the target in all directions when the quadrotor flies along the relocating trajectory.		
		
	We also conduct an outdoor experiment in a dense forest, as illustrated in Fig.\ref{pic:outdoor_experiment}. The target zooms through the massive trees. Strong winds throw up leaves and dust, making the self localization and map less credible. In spite of this, it turns out that the quadrotor still reaches $3m/s$ and closely follows the target in such a challenging environment. 
	
	From the experiments, we conclude that the proposed system maintains robustness in challenging tasks and present satisfactory tracking performance as well. More details are available in the attached video.

	\begin{table}
		\centering
		\caption{Comparison in the Sharp-Turning scenario.}
			\setlength{\tabcolsep}{6mm}
		\renewcommand\arraystretch{1.5}
		\label{tab:planner_cmp}
		\begin{tabular}{|c|c|c|} 
			\hline
			Scenario                    & Method   & Success Rate  \\ 
			\hline
			\multirow{2}{*}{Low Speed}  & Proposed & \bf 10/10         \\ 
			\cline{2-3}
			& Fast-Tracker~\cite{HZP2020tracker} & 8/10          \\ 
			\hline
			\multirow{2}{*}{High Speed} & Proposed & \bf 9/10          \\ 
			\cline{2-3}
			& Fast-Tracker~\cite{HZP2020tracker} & 3/10          \\
			\hline
		\end{tabular}
	\vspace{-1.2cm}
	\end{table}

  	\begin{figure}[t]
	\centering
	\includegraphics[width=1\linewidth]{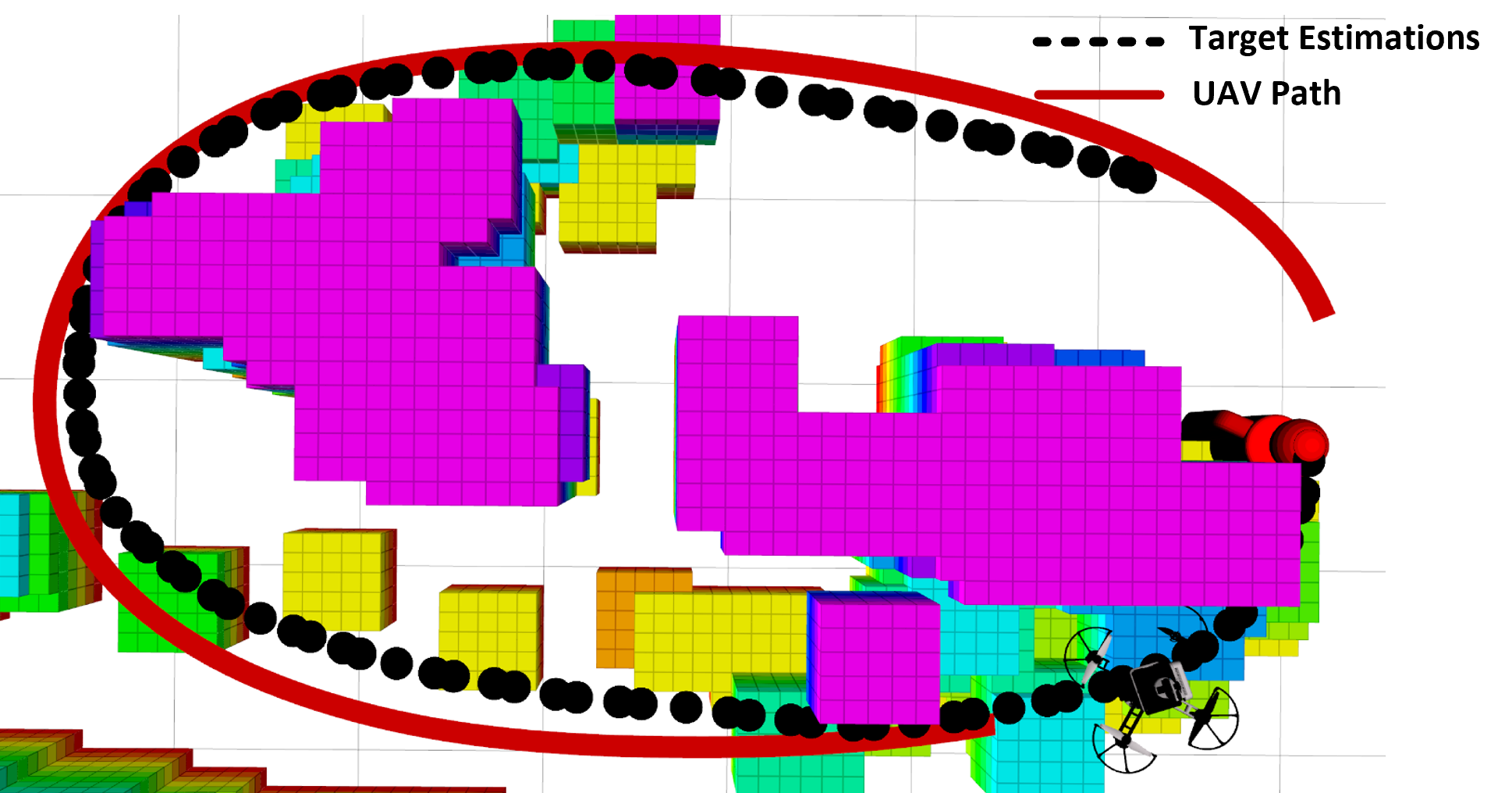}
	\captionsetup{font={small}}
	\caption{
		Visualization of the experimental scenario in benchmark comparisons. The sharp turning and massive obstacles are likely to cause occlusion, making this scenario challenging. 
	}
	\label{pic:benchmark_env}
	\vspace{-0.2cm}
\end{figure}

	\subsection{Benchmark Comparisons}	
	We compare the proposed system with Fast-Tracker\cite{HZP2020tracker}. We set a tracking scenario with sharp turnings, as illustrated in Fig.\ref{pic:benchmark_env}. When the target suddenly turns, the surrounding obstacles are likely to cause occlusion. 10 tests are conducted when the target moves in low speed ($0.85m/s$) and high speed ($1.5m/s$), respectively. A test fails when the quadrotor cannot follow the target all the way. Tab.\ref{tab:planner_cmp} reveals that even if the proposed system faces more difficult detection tasks than Fast-Tracker~\cite{HZP2020tracker}, it still surpasses \cite{HZP2020tracker} in the tracking performance. With active vision, the target is kept to be detected even when performing jerky motions. Furthermore, the occlusion-aware path searching method tends to search for grid paths with better visibility.

	\section{Conclusion}
	\label{sec:conclusion}
	In this paper, we improve our previous work Fast-Tracker\cite{HZP2020tracker} to apply it in human target tracking scenarios. We propose a lightweight and practical target detection and localization method utilizing monocular images. To avoid occlusion against obstacles, we implement active vision on a customized gimbal camera, together with an occlusion-aware mechanism in the motion planning module. Extensive real-world experiments confirm that the proposed approach can demonstrate real-time performance on an onboard platform and maintain robustness in complex environments. Furthermore, comparisons with Fast-Tracker reveal that the proposed active vision and occlusion-aware mechanism reinforce target perception capability and make the proposed system applicable to more challenging tracking scenarios.
 	
 	In the future, we will focus on the artistic decision-making in aerial tracking, such as the autonomous selection of shooting angles and relative motion patterns under different scene contexts.
	Moreover, we will try to apply the proposed approach to other types of UAVs so as to handle more challenging tracking scenarios. 
	\vspace{1.5cm}
	\bibliography{iros2021_pz}

\begin{thebibliography}{10}
\providecommand{\url}[1]{#1}
\csname url@rmstyle\endcsname
\providecommand{\newblock}{\relax}
\providecommand{\bibinfo}[2]{#2}
\providecommand\BIBentrySTDinterwordspacing{\spaceskip=0pt\relax}
\providecommand\BIBentryALTinterwordstretchfactor{4}
\providecommand\BIBentryALTinterwordspacing{\spaceskip=\fontdimen2\font plus
\BIBentryALTinterwordstretchfactor\fontdimen3\font minus
  \fontdimen4\font\relax}
\providecommand\BIBforeignlanguage[2]{{%
\expandafter\ifx\csname l@#1\endcsname\relax
\typeout{** WARNING: IEEEtran.bst: No hyphenation pattern has been}%
\typeout{** loaded for the language `#1'. Using the pattern for}%
\typeout{** the default language instead.}%
\else
\language=\csname l@#1\endcsname
\fi
#2}}

\bibitem{TN2017elli}
T.~{Nägeli}, J.~{Alonso-Mora}, A.~{Domahidi}, D.~{Rus}, and O.~{Hilliges},
  ``Real-time motion planning for aerial videography with dynamic obstacle
  avoidance and viewpoint optimization,'' \emph{IEEE Robotics and Automation
  Letters}, vol.~2, no.~3, pp. 1696--1703, 2017.

\bibitem{Bonatti2019JFR}
R.~Bonatti, W.~Wang, C.~Ho, A.~Ahuja, M.~Gschwindt, E.~Camci, E.~Kayacan,
  S.~Choudhury, and S.~Scherer, ``{Autonomous aerial cinematography in
  unstructured environments with learned artistic decision-making},''
  \emph{Journal of Field Robotics}, vol.~37, no.~4, pp. 606--641, 2020.

\bibitem{BJ2020ICRA}
B.~{Jeon}, Y.~{Lee}, and H.~J. {Kim}, ``Integrated motion planner for real-time
  aerial videography with a drone in a dense environment,'' in \emph{2020 IEEE
  International Conference on Robotics and Automation (ICRA)}, 2020, pp.
  1243--1249.

\bibitem{JC2016tracking}
J.~{Chen}, T.~{Liu}, and S.~{Shen}, ``Tracking a moving target in cluttered
  environments using a quadrotor,'' in \emph{2016 IEEE/RSJ International
  Conference on Intelligent Robots and Systems (IROS)}, 2016, pp. 446--453.

\bibitem{HZP2020tracker}
Z.~Han, R.~Zhang, N.~Pan, C.~Xu, and F.~Gao, ``Fast-tracker: A robust aerial
  system for tracking agile target in cluttered environments,'' \emph{arXiv
  preprint arXiv:2011.03968}, 2020.

\bibitem{JW2013noc}
J.~{Kim} and D.~H. {Shim}, ``A vision-based target tracking control system of a
  quadrotor by using a tablet computer,'' in \emph{2013 International
  Conference on Unmanned Aircraft Systems (ICUAS)}, 2013, pp. 1165--1172.

\bibitem{AG2014noc}
A.~G. {Kendall}, N.~N. {Salvapantula}, and K.~A. {Stol}, ``On-board object
  tracking control of a quadcopter with monocular vision,'' in \emph{2014
  International Conference on Unmanned Aircraft Systems (ICUAS)}, 2014, pp.
  404--411.

\bibitem{HC2017noc}
H.~{Cheng}, L.~{Lin}, Z.~{Zheng}, Y.~{Guan}, and Z.~{Liu}, ``An autonomous
  vision-based target tracking system for rotorcraft unmanned aerial
  vehicles,'' in \emph{2017 IEEE/RSJ International Conference on Intelligent
  Robots and Systems (IROS)}, 2017, pp. 1732--1738.

\bibitem{Huang2018Openpose}
C.~{Huang}, Z.~{Yang}, Y.~{Kong}, P.~{Chen}, X.~{Yang}, and K.~{Cheng},
  ``Through-the-lens drone filming,'' in \emph{2018 IEEE/RSJ International
  Conference on Intelligent Robots and Systems (IROS)}, 2018, pp. 4692--4699.

\bibitem{MobileNet}
A.~G. Howard, M.~Zhu, B.~Chen, D.~Kalenichenko, W.~Wang, T.~Weyand,
  M.~Andreetto, and H.~Adam, ``Mobilenets: Efficient convolutional neural
  networks for mobile vision applications,'' \emph{arXiv preprint
  arXiv:1704.04861}.

\bibitem{ren2015faster}
S.~Ren, K.~He, R.~Girshick, and J.~Sun, ``Faster {R-CNN}: Towards real-time
  object detection with region proposal networks,'' in \emph{Neural Information
  Processing Systems ({NIPS})}, 2015.

\bibitem{Galvane2018}
Q.~Galvane, C.~Lino, M.~Christie, J.~Fleureau, F.~Servant, F.-l. Tariolle, and
  P.~Guillotel, ``Directing cinematographic drones,'' \emph{ACM Trans. Graph.},
  vol.~37, no.~3, July 2018.

\bibitem{JOUBERT2016}
N.~{Joubert}, J.~L. {E}, D.~B. {Goldman}, F.~{Berthouzoz}, M.~{Roberts}, J.~A.
  {Landay}, and P.~{Hanrahan}, ``{Towards a Drone Cinematographer: Guiding
  Quadrotor Cameras using Visual Composition Principles},'' \emph{arXiv
  preprint arXiv:1712.04216}, 2016.

\bibitem{Openpose}
Z.~{Cao}, G.~{Hidalgo Martinez}, T.~{Simon}, S.~{Wei}, and Y.~A. {Sheikh},
  ``Openpose: Realtime multi-person 2d pose estimation using part affinity
  fields,'' \emph{IEEE Transactions on Pattern Analysis and Machine
  Intelligence}, 2019.

\bibitem{gao2018online}
F.~{Gao}, W.~{Wu}, Y.~{Lin}, and S.~{Shen}, ``Online safe trajectory generation
  for quadrotors using fast marching method and bernstein basis polynomial,''
  in \emph{2018 IEEE International Conference on Robotics and Automation
  (ICRA)}, 2018, pp. 344--351.

\bibitem{DD2008HybridAstar}
D.~Dolgov, S.~Thrun, M.~Montemerlo, and J.~Diebel, ``{Practical search
  techniques in path planning for autonomous driving},'' \emph{International
  Symposium on Combinatorial Search, SoCS 2008}, 2008.

\bibitem{zhou2019robust}
B.~{Zhou}, F.~{Gao}, L.~{Wang}, C.~{Liu}, and S.~{Shen}, ``Robust and efficient
  quadrotor trajectory generation for fast autonomous flight,'' \emph{IEEE
  Robotics and Automation Letters}, vol.~4, no.~4, pp. 3529--3536, 2019.

\bibitem{wang2020DDP}
Z.~Wang, H.~Ye, C.~Xu, and F.~Gao, ``Generating large-scale trajectories
  efficiently using double descriptions of polynomials,'' \emph{arXiv preprint
  arXiv:2011.02662}, 2020.

\bibitem{qin2017vins}
T.~Qin, P.~Li, and S.~Shen, ``Vins-mono: A robust and versatile monocular
  visual-inertial state estimator,'' \emph{IEEE Transactions on Robotics},
  vol.~34, no.~4, pp. 1004--1020, 2018.

\end{thebibliography}
\end{document}